# Vibration fault detection in wind turbines based on normal behaviour models without feature engineering

S. Jonas[1], D. Anagnostos[2], B. Brodbeck[2], A. Meyer[1]

[1] Bern University of Applied Sciences, Quellgasse 21, 2501 Biel, Switzerland
[2] WinJi AG, Badenerstrasse 808, 8048 Zurich, Switzerland

**Abstract.** Most wind turbines are remotely monitored 24/7 to allow for an early detection of operation problems and developing damage. We present a new fault detection method for vibration-monitored drivetrains that does not require any feature engineering. Our method relies on a simple model architecture to enable a straightforward implementation in practice. We propose to apply convolutional autoencoders for identifying and extracting the most relevant features from the half spectrum in an automated manner, saving time and effort. Thereby, a spectral model of the normal vibration response is learnt for the monitored component from past measurements. We demonstrate that the model can successfully distinguish damaged from healthy components and detect a damaged generator bearing and damaged gearbox parts from their vibration responses. Using measurements from commercial wind turbines and a test rig, we show that vibration-based fault detection in wind turbine drivetrains can be performed without the usual upfront definition of spectral features. Another advantage of the presented method is that the entire half spectrum is monitored instead of the usual focus on monitoring individual frequencies and harmonics.

## 1. Introduction

Wind energy is becoming an essential pillar of the energy mix, helping to decarbonize the energy system and to improve energy independence and security of supply in many countries (IEA, 2021a,b). Most commercial wind farms today are remotely monitored around the clock to keep the operation and maintenance costs low. The continuous monitoring allows for an early detection of potential operation problems and facilitates proactive condition-based maintenance. Gearboxes and generators are of particular interest in the monitoring of wind turbines (WT) because they are especially costly to replace and associated replacement work tends to entail long downtimes (Tavner et al., 2006; Faulstich et al., 2011; Crabtree et al., 2015; Carroll et al., 2016). Thus, more and more wind turbine drive trains are monitored by accelerometers. This makes it possible to derive and track the vibration spectra of critical components such as gearbox parts and generator bearings.

Previous studies have proposed various methods and features for vibration-based fault detection in the drive trains of wind turbines. The proposed frequency-domain methods include the monitoring and analysis of particular frequencies, their harmonics, sidebands and the signal envelope. In the time domain, the existing methods focus on analyzing the amplitudes of the vibration time series and tracking statistical properties of the vibration response distribution. We refer to Salameh et al., 2018 and Barszcz, 2019 for comprehensive reviews of the state-of-the-art methods of vibration-based fault detection in wind turbine drive trains.

The existing methods require the upfront development and extraction of component-specific features from the accelerometer measurements by condition monitoring engineers. The upfront definition of spectral features for the monitoring constitutes a major time investment before commissioning and operating the turbine. For instance, in-depth information about the respective gearbox design and composition are needed to this end. The characteristic frequencies of the monitored components need to be determined if the corresponding spectral lines are supposed to be tracked. Collecting this spectral information for every monitored component in every monitored wind turbine and model constitutes a major effort. The extracted spectral features tend to be turbine- and component-specific. Therefore, they can generally not be reused for new wind turbine types or when components have been updated. Moreover, feature engineering can result in more and in less effective features for fault detection. For example, in a case study by Koukoura et al., 2019, spectral line and cepstrum analyses resulted in higher fault detection accuracies than time synchronous averaging and spectral kurtosis methods. Thus, depending on the chosen feature engineering approach, one may end up with more or less satisfactory fault detection accuracies.

Recently, first studies have proposed the application of fault detection methods that do not require any feature engineering. Li et al., 2018 demonstrated the application of autoencoders for detecting blade damages in wind turbines based on the blade stress and strain signals obtained from strain gauge sensors. In a similar application, Yang et al., 2021 made use of a convolutional autoencoder for detecting blade damages. To this end, they employed data from the wind turbine's supervisory control and data acquisition (SCADA) system to identify changes in the dynamics of the blade system. Li et al., 2021 applied an autoencoder for the transfer learning of fault diagnosis tasks on SCADA and failure status datasets. They investigated different fault types and focused on the transfer of fault diagnostics models to target turbines with little available SCADA data. When vibration sensors are unavailable, SCADA-based modelling of the turbine's normal behavior can enable the detection of operation faults in drive train components (e.g., Zaher et al., 2009; Schlechtingen et al., 2013; Tautz-Weinert et al., 2017; Meyer et al., 2020, 2021).

The goal of our study is to develop and demonstrate a fault detection method for vibration-monitored drivetrain components in wind turbines that does not require any feature engineering but learns the characteristic spectral features of the monitored components without any human assistance from the entire half spectrum. The method should enable simple model architectures to facilitate its adoption by practitioners. At the same time, it should neither be restricted to monitoring specific frequencies nor should it require a comprehensive set of gearbox- and fault-type-specific observations. The latter would be needed for fault detection methods based on supervised machine learning but in practice, such datasets are usually not available or not accessible to wind turbine operators.

To achieve this goal, the present study proposes and demonstrates the application of convolutional autoencoders for the feature learning, extraction and early detection of faults in vibration-monitored wind turbine gearboxes and generators. This study is the first to propose spectral normal behaviour models constructed without feature engineering for the purpose of vibration-based fault detection in wind turbine drive trains, to the best of our knowledge.

In this study, we also propose and investigate fault detection based on non-convolutional autoencoders. Moreover, we compare the performance of reconstruction-error-based gearbox and generator health indices to health indices derived by one-class classification with an isolation forest model.

This paper is organized as follows. Section 2 describes our new fault detection methods. Section 3 introduces the gearbox and generator datasets and models used for demonstrating the methods. The results of our study are discussed in section 4, and our conclusions follow in section 5.

## 2. Fault detection algorithms

Autoencoders are artificial neural networks that are trained to compress and then reconstruct their input (e.g., Bourlard et al., 1988; Kramer, 1991; Goodfellow et al., 2016). Autoencoders are autoassociative in that they are trained to duplicate their input as their output and to minimize the resulting reconstruction error. In the training process, the reconstruction error is minimized by adapting the autoencoder's network weights to the training data in an iterative manner. Autoencoders consist of encoding and decoding hidden layers. The encoding layers learn a lower dimensional representation of the training data which is subsequently decompressed by the decoding layers to reconstruct an output highly similar to the input with the same dimensionality as the input. The dimensionality reduction of the input data has a similar compression effect as a principal component analysis. The proposed convolutional autoencoders reduce the dimension of the spectral feature space and replace human feature engineering. Convolutional autoencoders are convolutional neural networks with an encoder-decoder architecture. They are particularly suited for image data compression and decompression because of the feature learning and image reconstruction capabilities of the convolutional layers. Autoencoders can be used to detect anomalous input data instances based on the large reconstruction errors they make when trying to reconstruct those instances from their compressed representations. Therefore, convolutional autoencoders have been successful in image and video-based anomaly detection tasks (e.g., Ribeiro et al., 2018; Duman et al., 2019).

In this study, we propose to apply convolutional autoencoders to learn and extract spectral features from the accelerometer measurements and to use the autoencoder reconstruction errors as an indicator (health index) of unusual operation behaviour and potential faults of the monitored components. Alternatively, the extracted features can be input to a one-class classification algorithm, such as an isolation forest model, to construct a health index of the monitored component, as demonstrated in our case study. We present two fault detection methods below that rely on two steps: (1) automated data-driven feature learning and extraction by autoencoders to model the normal vibration responses of the monitored component, and (2) subsequent fault detection based on deviations from the spectral normal behaviour model learnt in step (1). The two fault detection methods may be used individually or in combination. We demonstrate and discuss them in a case study in sections 3 and 4.

**2.1 Fault detection based on autoencoder reconstruction losses**

As shown in Figure 1, an autoencoder is trained with spectrogram segments constructed from the accelerometer measurements. The autoencoder's reconstruction loss is used as an anomaly score (health index) in this case. If the autoencoder has been trained only on vibration spectra from components unaffected by faults, then the reconstruction error of spectra from fault-affected components is expected to be higher than that of unaffected components. Thus, the higher the reconstruction loss of an input segment, the more likely its underlying spectrogram is to indicate a fault in the monitored component. In this approach, the autoencoder should be trained on data from healthy components so it learns to reconstruct spectra that reflect the normal operation behaviour of the monitored component. Thus, the autoencoder learns a normal behaviour model of the vibration responses of the component. The underlying idea of this unsupervised fault detection approach is similar to that of the normal behaviour models in SCADA-based condition monitoring (Zaher et al., 2009; Schlechtingen et al., 2013; Tautz-Weinert et al., 2017; Meyer et al., 2020, 2021). Training, validation and test sets are created from the spectrograms constructed from the vibration responses of each monitored component. The mean absolute reconstruction error is calculated for each segment by comparing its autoencoder reconstruction to the original input. To identify fault-affected spectrograms, we define a loss threshold $T$. When this threshold is exceeded, the component associated with the spectrogram is considered to be fault-affected.

Formally, an autoencoder consists of encoder network $\phi_e$ and decoder network $\phi_d$, parameterized by $\Theta_e$ and $\Theta_d$ respectively, derived by minimizing the reconstruction loss during training. The health index $h$ assigned to a spectrogram segment $X_k$ recorded by accelerometer $s$ is obtained by calculating the mean absolute error $\mathcal{L}_{MAE,AE_s}$ between the original input $X_k$ and its reconstruction $\widehat{X_k}$ by an autoencoder $AE_s$ that was trained on only healthy segments from accelerometer $s$. The classification as fault-affected is based on whether the health index $h$ exceeds a predefined threshold $T_s$. This threshold will ensure that normal variability in the spectrogram will not be flagged as fault conditions whereas unusual spectral changes will be detected.

$$\widehat{X_{k,s}} = \phi_{d,s}\big(\phi_{e,s}(X_{k,s};\Theta_{e,s});\Theta_{d,s}\big)$$

$$h(X_{k,s}) = \mathcal{L}_{MAE,AE_s}(X_{k,s}, \widehat{X_{k,s}})$$

$$anomaly(X_{k,s}) = \text{true}, \quad \text{if } h(X_{k,s}) > T$$

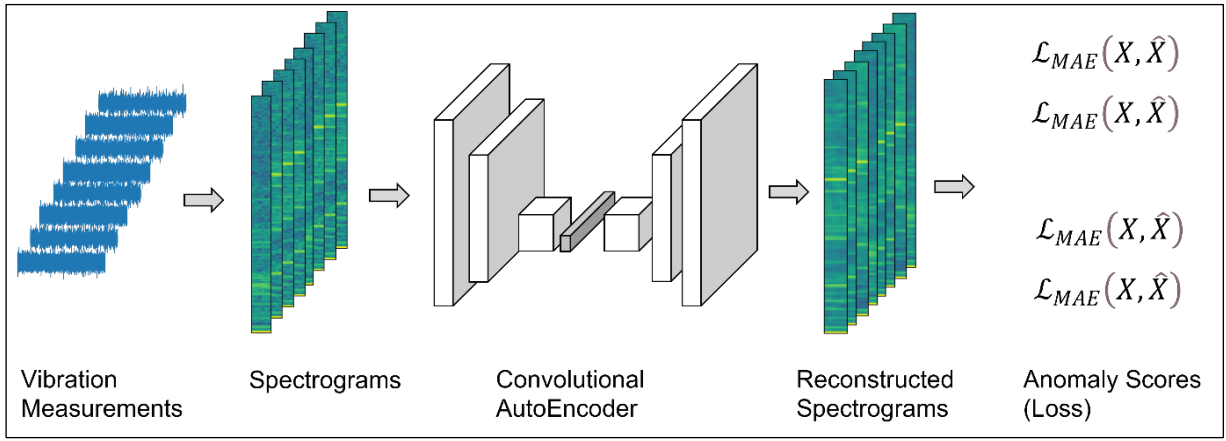

**Figure 1.** An illustration of the fault detection approach based on autoencoder reconstruction errors. First, the vibration measurements are converted to spectrogram segments. The segments are given as input to a convolutional autoencoder which returns reconstructed spectrograms as output. The mean absolute reconstruction errors are calculated with respect to the original inputs. These errors are used as health indices to distinguish healthy and fault-affected components.

### 2.2 Fault detection based on isolation forests

The second fault detection approach presented in this study combines the encoding part of a convolutional autoencoder with one-class classification by an isolation forest model, as shown in Figure 2. As in the first approach, a convolutional autoencoder is trained with the goal to reconstruct spectrogram segments. In contrast to the first approach, however, the autoencoder will now only be used to extract a compressed representation of the most important features from the input spectrogram. In a second step, the extracted features serve as input to an isolation forest model (Liu et al., 2008). The features are taken from the output of the last encoding layer of the autoencoder architecture. In this study, we call this layer the bottleneck layer.

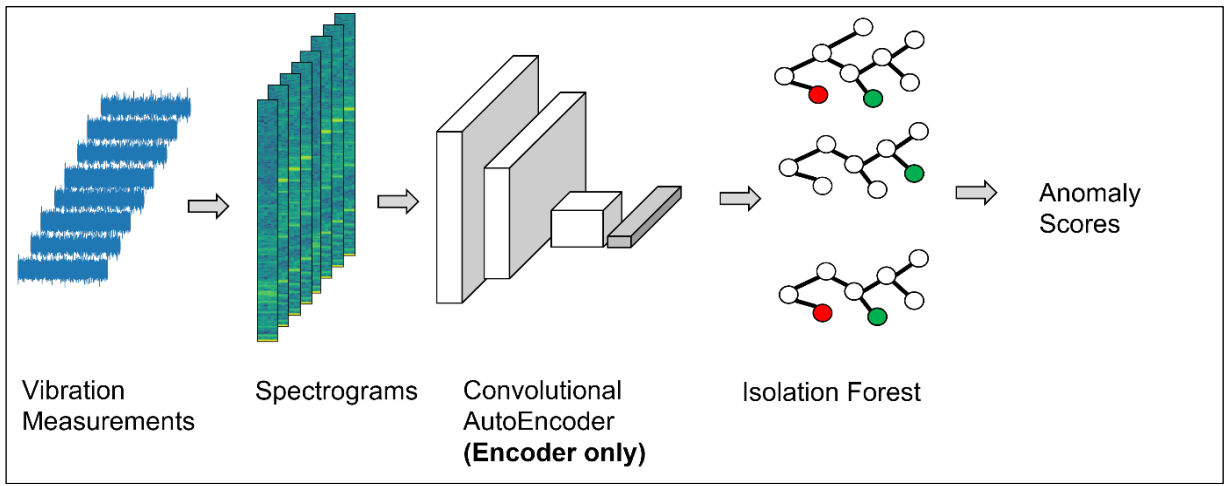

**Figure 2.** An illustration of the fault detection approach based on isolation forests. First, the vibration measurements are converted to spectrogram segments. The segments are given as input to the encoding part of a convolutional autoencoder whose last layer outputs an encoded representation of the segment features (shaded layer). Then, an isolation forest model computes an anomaly score for each segment. The score is used as a health index to distinguish healthy and fault-affected components.

The isolation forest model detects an anomalous spectrogram based on how easily the spectrogram can be isolated from the remaining spectrograms in the decision trees constituting the forest. In this way, the isolation forest model distinguishes anomalous spectrograms from spectrograms featuring normal operation behaviour. A separate isolation forest model is created for each monitored component and accelerometer. The inputs to the forest are the feature vectors extracted from the bottleneck layer of the previously trained autoencoder. In a final step, the model evaluates all segments from the training, validation, and test sets, and outputs an anomaly score. We need to define a threshold to discriminate spectrograms that reflect normal operation behaviour from spectrograms of fault-affected components. This threshold is set to 0.5 in accordance with Liu et al., 2008.

More formally, for a spectrogram segment $X_k$ recorded by accelerometer $s$, the trained isolation forest model $IF$ for $s$ outputs the health index $h$ based on the feature vector $z(X_k)$ output from the encoder network $\phi_{e,s}$ of autoencoder $AE_s$:

$$z(X_{k,s}) = \phi_{e,s}(X_{k,s}; \Theta_{e,s})$$

$$h(X_{k,s}) = IF_s\left(z(X_{k,s})\right)$$

$$anomaly(X_{k,s}) = \text{true}, \quad \text{if } h(X_{k,s}) > 0.5$$

## 3. Case study

The proposed fault detection methods have been demonstrated in four wind turbine drive trains. First, we tested the approach in two commercially operated wind turbines which we call wind turbine 1 (WT1) and wind turbine 2 (WT2) in this study. We knew from the operators that WT1 exhibited no known damages nor functional impairments, whereas WT2 may have suffered damage in a drive train component. Our goal was to test the proposed fault detection methods by detecting if any of the monitored components of WT2 are affected by faults based on their vibration responses, and to identify the fault-affected components if any.

In addition to the commercial wind turbines, we also tested the performance of our fault detection methods with test rig measurements from the National Renewable Energy Laboratory (NREL).

### 3.1 Commercially operated wind turbines

Accelerometer measurements from the drive trains of two commercially operated multi-MW wind turbines were analyzed. The two onshore turbines were identical in design and composition with a nominal power of 2 MW. Table 1 provides further specifications of the turbines.

In each turbine, eight accelerometers monitored the gearbox, generator, main shaft bearing and the main frame, as detailed in Table 2. Figure 3 illustrates the monitored components and the locations of the accelerometers. The measurements were taken at a sampling frequency of 25.6 kHz in time slices of 2 seconds each over a period of three months. All time slice measurements were taken at the same generator load of 1345 ± 3 rpm. The constant load enables a direct comparison of the spectra across the full measurement period of three months and among the two turbines. In total, the dataset contained 562 time slice recordings from WT1 and 672 recordings from WT2.

| Quantity | Value |
|---|---|
| Rotor diameter | 114 m |
| Nominal power | 2.1 MW |
| Type | Variable-speed horizontal-axis pitch-controlled |
| Deployment | onshore |
| Gearbox | Three stage |

**Table 1.** Technical specifications of the two commercial wind turbines.

| Accelerometer ID | Monitored component and direction |
|---|---|
| S1 | Main frame, axial |
| S2 | Gearbox, rotor side, axial |
| S3 | Main frame, radial |
| S4 | Gearbox, generator side, radial |
| S5 | Main shaft bearing, radial |
| S6 | Gearbox, generator side, axial |
| S7 | Generator, coupling side, axial |
| S8 | Generator, coupling side, radial |

**Table 2.** The monitored components and measured acceleration directions in the two wind turbines.

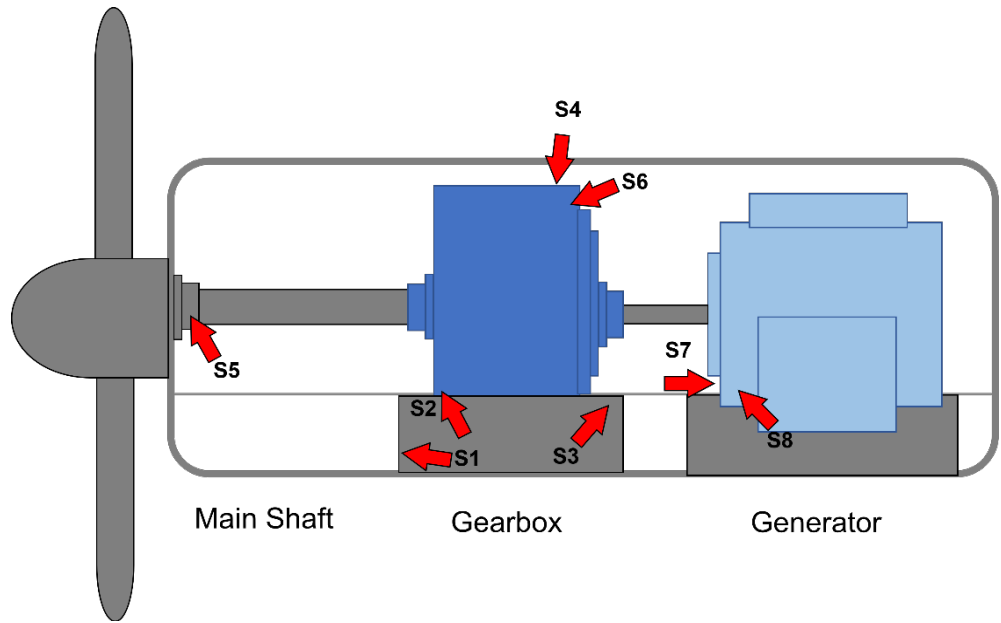

**Figure 3.** Sketch of the drive train of the two wind turbines. The accelerometers S1-S8 are indicated in red.

### 3.2 Test rig measurements

In addition to the commercial wind turbines, we tested our fault detection methods with accelerometer measurements from the drive trains of two wind turbines operated on an NREL test rig. The turbines were two 750 kW machines identical in construction. The gearboxes provided a transmission of 1:81.5 and were composed of a low-speed planetary stage and two parallel moderate and high-speed stages. One of the gearboxes showed no damage or functional impairments. The other gearbox had suffered an oil loss and resulting damages in multiple of its components. The vibration measurements were taken from multiple components of the gearboxes as part of the Gearbox Condition Monitoring Round Robin study (Sheng, 2012). The measurements were conducted for ten minutes and under constant load at both gearboxes with 22 rpm of the low-speed shaft and 1800 rpm of the high-speed shaft. The accelerometers measured at a 40 kHz sampling frequency. A comprehensive description of the gearbox, the sensing system and test environment are provided by Sheng, 2012 and Musial et al., 2000.

The vibration measurements analysed in this study were obtained from components which exhibited different damage levels in the gearbox that had suffered the oil loss, such as scuffing damage. Their counterparts in the healthy gearbox have been examined for damage as part of the Round Robin study (Sheng, 2012) and no damage was found in them. The accelerometers monitored the vibrations in the following locations in the damaged and in the healthy gearbox:
1. in a bottom-facing position at the ring gear,
2. at the bearing of the high-speed shaft, and
3. at the bearing of the low-speed shaft.

All accelerometers measured the accelerations in the respective radial direction.

## 4. Results and discussion

To convert the accelerometer measurements into spectrograms, we applied a short-time Fourier transform (STFT; Allen, 1977) with window sizes of 250 ms and an overlap of 100 ms. Fault detection is performed based on spectrogram segments of 1 second duration and considering frequencies up to 1000 Hz. These parameters were chosen so as to enable sufficient temporal evolution and frequency resolution of the signal and at the same time allow for large enough training, validation and test sets. Our results are robust against modifications of these parameters. Prior to the model training, we applied a log-transformation to each spectrogram and a min-max normalization such that all data points fall in the range of [0,1]. Figure 4 shows examples of the resulting spectrograms for the components monitored by the accelerometers S1-S8 in both wind turbines.

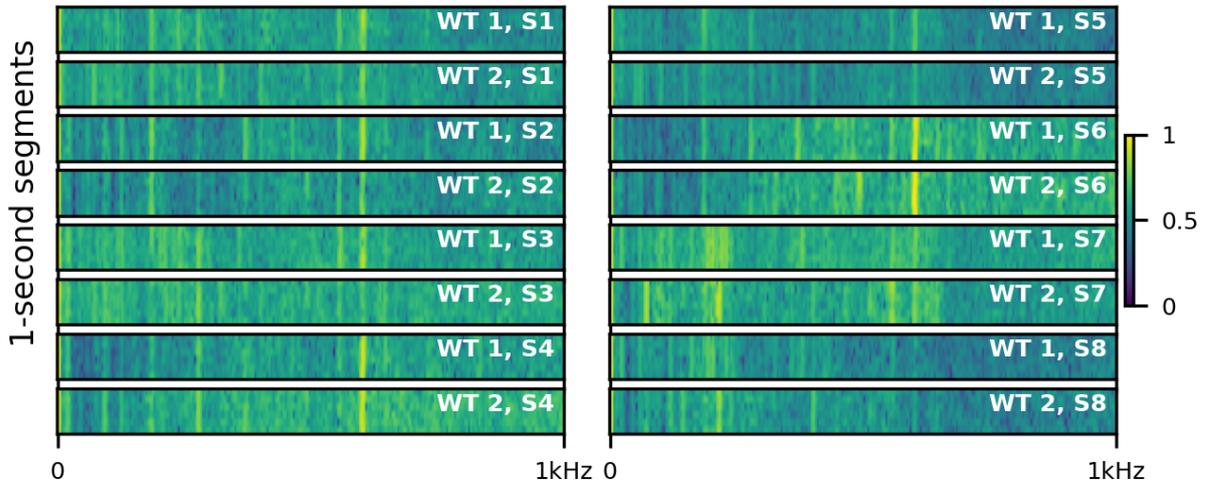

**Figure 4.** Sample spectrograms of WT1 and WT2 from the components monitored by the accelerometers S1-S8

**Dataset splits.** For our case study, we created a separate model for each accelerometer. We split the recordings of WT1 into a 70% training set, a 15% validation set and a 15% test set for each accelerometer. A sliding window was applied to extract multiple 1-second segments from each 2-second recording. The resulting segments of the training set were used to train the autoencoders and the isolation forests, while the validation set was used for the model selection and early stopping. All data of WT2 served as an additional separate test set, so we can obtain fault notifications for each component.

**Model selection.** We performed a preliminary model search using the Hyperband hyperparameter optimization algorithm (Li et al., 2017) in order to find an optimal convolutional autoencoder architecture. We evaluated architectures consisting of only convolutional and pooling layers which resulted in a bottleneck size of 128 units (the flattened size of the last encoder layer). This dimension requirement was set beforehand by us. Choosing a bottleneck size significantly smaller than the input size forces the network to learn a compressed representation of the inputs. This compressed representation serves as input to the isolation forest. In terms of configurations, we evaluated models with up to 5 encoding layers with either 4, 8, or 16 feature maps, with kernel sizes (height, width) of either (1,3), (1,5), (3,3), (3,5), or (5,5) and with varying learning rates between 3e-2 and 1e-4. The decoder architecture was always symmetrical to the encoder. The network weights were optimized using the Adaptive Moment Estimation (Adam) optimization algorithm (Kingma et al., 2015), minimizing the mean absolute error (MAE) between the reconstruction output and the original input. Further, we applied an early stopping mechanism to stop training when the validation loss had not improved within 15 epochs.

This hyperparameter optimization was performed only on the training- and validation set of accelerometer S1. The Hyperband search algorithm resulted in the best performing convolutional architecture ("conv-AE") outlined in Table 3. Additionally, we compared the reconstruction performance to a minimal dense architecture ("dense-AE") described in Table 3, i.e., to the smallest possible model configuration in terms of parameters with at least one hidden fully-connected layer and the same bottleneck size.

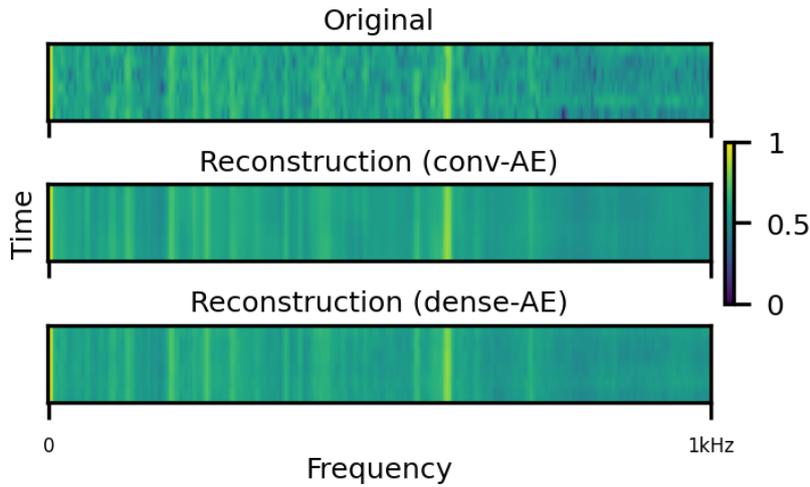

**Figure 5.** An example of the reconstruction capabilities of the two considered autoencoder models. An original spectrogram segment (top) is given as input to an autoencoder which outputs a reconstruction. A convolutional and a dense autoencoder were adapted to reconstruct a spectrogram derived from accelerometer S1 in this illustration. The reconstruction by the dense autoencoder (bottom) suggests that this model also tries to fit more of the noise in the spectrogram, which is consistent with the overfitting observed with the dense autoencoder.

Both the convolutional and the dense model architectures evaluated by us are capable of reconstructing visually similar spectrogram segments after finished training, as shown in Figures 4 and 5. While the training loss was lower using the dense autoencoder model (3.55e-2) compared to the convolutional autoencoder (3.63e-2), the validation loss was higher (3.82e-2) compared to the convolutional autoencoder model (3.70e-2), indicating a poorer reconstruction performance of the dense autoencoder model on the same unseen dataset. We attribute the divergence of the training and validation losses observed in the case of the dense autoencoder to its overfitting on the training set due to the dense autoencoder's large number of parameters. Our case study showed that the convolution-based architecture can achieve a better reconstruction performance on unseen data with a comparably small number of parameters, specifically with only 2.5% of the dense autoencoder model's number of parameters. At the same time, the convolutional autoencoder maintained a good correspondence between training and validation losses. This suggests that the convolutional autoencoder network learns more generalizable features and is less prone to overfitting in this application. In addition, the dense autoencoder has a large number of parameters so its training can quickly become computationally expensive when considering multiple models or larger datasets with even more components and wind turbines. Based on these results, we determine that the convolutional architecture is better suited for our task. Consequently, we proceeded by using this configuration for all further experiments.

**Fault detection based on autoencoder reconstruction losses.** We train a separate convolutional autoencoder network with the conv-AE configuration for each accelerometer using 70% of the measurements of the monitored healthy component from WT1 as training set and 15% as validation set. The networks are trained with the same procedure as outlined in the model selection experiment. During the training, all eight autoencoders achieved similar performances in terms of their validation losses (3.48e-2 – 4.68e-2). We evaluate the reconstruction errors obtained from the reconstructions of the segments in the training, validation and test sets of WT1 and the test data from WT2. A threshold was determined based on the training errors, and if it was exceeded, we considered a segment as anomalous.

| Model | Architecture |
|---|---|
| Convolutional autoencoder (conv-AE) | • Input: 6 x 251 x 1 (height x width x channels), zero-padded to 8 x 256 x 1<br>• Encoder:<br>Block 1: Conv. Layer (16f, 1x3, elu), MP (1,2)<br>Block 2: Conv. Layer (16f, 3x3, elu), MP (2,2)<br>Block 3: Conv. Layer (16f, 1x3, elu), MP (2,2)<br>Block 4: Conv. Layer (16f, 1x5, elu), MP (1,2)<br>Block 5: Conv. Layer (8f, 1x5, elu), MP (1,2)<br>Bottleneck Layer: Flatten (2H x 8W x 8f to 128 units)<br>• Decoder:<br>Reshape: (128 units to 2H x 8W x 8f)<br>Block 1: Upsampling (1,2), Transposed Conv. Layer (8f, 1x5, elu)<br>Block 2: Upsampling (1,2), Transposed Conv. Layer (16f, 1x5, elu)<br>Block 3: Upsampling (2,2), Transposed Conv. Layer (16f, 1x3, elu)<br>Block 4: Upsampling (2,2), Transposed Conv. Layer (16f, 3x3, elu)<br>Block 5: Upsampling (1,2), Transposed Conv. Layer (16f, 1x3, elu)<br>• Reconstruction:<br>Cropping Layer: (8H x 251W x 16f to 6H x 251W x 16f)<br>Conv Layer: (1f, 5x5, sigmoid)<br><br>Number of parameters: 10'385 |
| Dense autoencoder (dense-AE) | • Input: 6 x 251 x 1 (height x width x channels) flattened to 1506 units<br>• Encoder:<br>Hidden Layer: FC Layer (128 units, elu)<br>Bottleneck Layer: FC Layer (128 units, elu)<br>• Decoder:<br>Hidden Layer: FC Layer (128 units, elu)<br>• Reconstruction Layer: FC Layer (1506 units, sigmoid)<br>Output reshaped back to 6 x 251 x 1<br><br>Number of parameters: 420'194 |
| Isolation Forest | Number of trees: 100<br>Contamination: 0.0001 |

**Table 3.** Specifications of the models used in our work. The abbreviations are "FC Layer": fully-connected layer (number of units, activation function); "Conv. Layer": convolutional layer (number of feature maps, kernel size in height x width, activation function); "MP": max-pooling layer (height x width); "elu": exponential linear unit.

Our autoencoder-based fault detection method finds increased reconstruction errors for the spectrograms derived from accelerometers S7 and S8, as shown in Figures 6 and 7. These sensors monitor the generator component coupling to the gearbox, as shown in Figure 3. The health index in Figure 7 displays the evolution of the health of each monitored component and the degree of anomaly of the component's vibrational response. Based on the sensors from WT1, we estimated that an appropriate threshold value for bounding the reconstruction errors of normal-vibration-response spectrograms is around 0.6. To estimate this threshold more accurately, a larger number of fault instances would be needed. Based on the health index, we define a custom rule for notifying the wind turbine operators when a certain number of anomalies were detected within a given timeframe. Specifically, a fault alarm was generated in our case study if the threshold was exceeded three times in a row, as shown by the shaded areas in Figure 7. No alarm was triggered for the components associated with S1-S6. Our findings indicate unusual and persistent spectral changes likely to result from fault-affected vibration responses of the monitored generator. We confirmed this finding by investigating the logs of the affected WT2. The logs specified that WT2 had suffered incipient generator damage without further detailing the type of damage. Thus, we could confirm that our proposed fault detection method is sufficiently sensitive to detect incipient generator damage from accelerometer measurements in commercial wind turbines. Note that we arrived at our diagnosis without any feature engineering but rather by letting the autoencoder itself learn what spectrograms look like in the normal operation of healthy components.

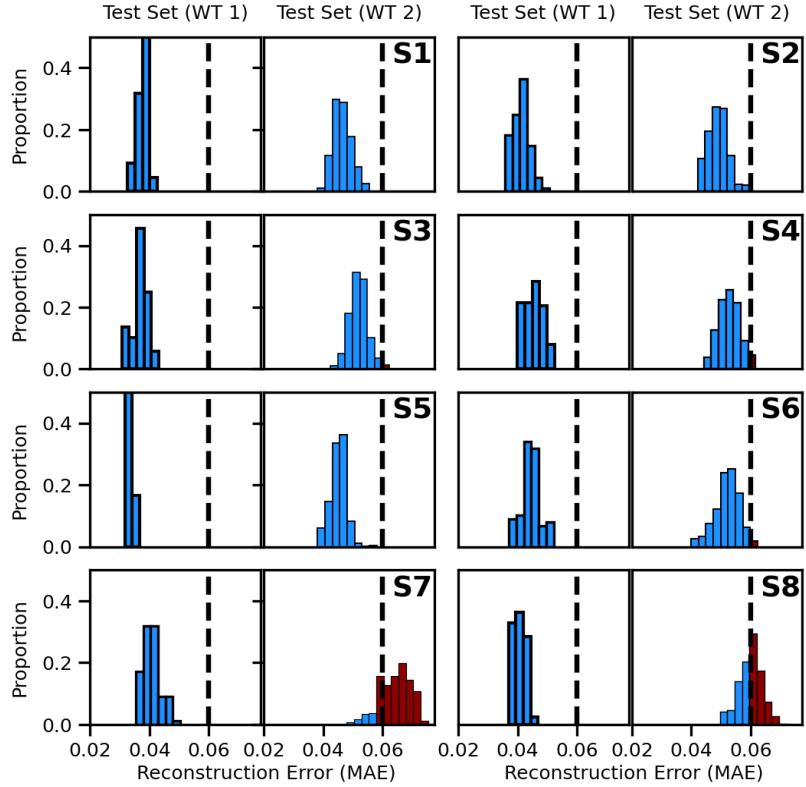

**Figure 6.** Fault detection based on spectrogram reconstruction. Reconstruction errors for accelerometers S1-S8 after reconstruction with a convolutional autoencoder. The spectrograms from the test set of WT1 are specified as healthy based on their comparatively low reconstruction errors, which is consistent with the logs of WT1. In contrast, the spectrograms of accelerometers S7 and S8 of WT2 indicate damage conditions of the generator. This finding was confirmed by the logs of WT2.

**Fault detection based on isolation forests.** We applied the previously trained autoencoders for the isolation-forest-based fault detection by using only their encoder parts. Each encoder part outputs a compressed feature representation in a vector of size 128 (the bottleneck layer output) which is then input to the isolation forests. For each accelerometer, we trained an isolation forest (IF) using the extracted feature vectors of spectrogram segments from the training set.

Each isolation forest was evaluated on unseen spectrograms from the test set of healthy WT1 and with spectrograms from WT2, as shown in Figure 8. The isolation forest models also provide an anomaly score (health index) for each spectrogram segment. Across all eight evaluated models, the isolation forests have consistently and correctly classified the WT1 test sets as healthy just like the reconstruction-error-based method, as shown in Figures 6 and 8. When evaluating the spectrograms from WT2, the isolation forest models assigned elevated anomaly scores to the generator (accelerometers S7, S8 in Figures 8-9). Figure 9 displays the health index for all monitored components. As shown in Figures 8 and 9, the fault detection based on isolation forests results in larger anomaly scores for the spectrograms derived from the accelerometers S7 and S8. This indicates a fault in the generator, which is consistent with the results of our reconstruction-error-based fault detection method.

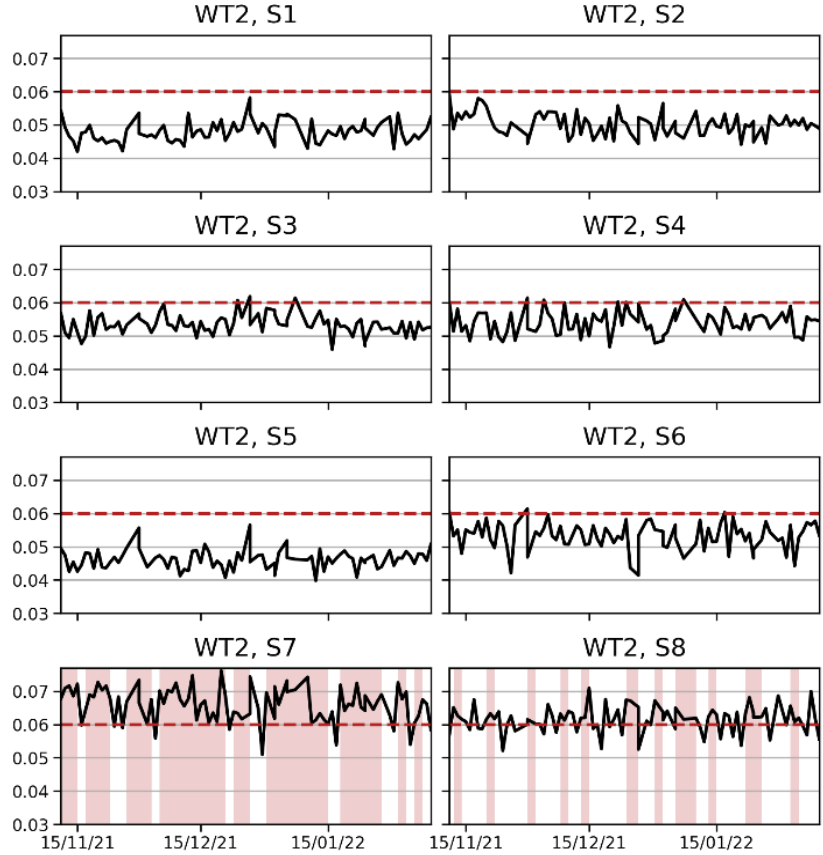

**Figure 7.** Fault detection based on spectrogram reconstruction. The reconstruction errors are considered health indices of the monitored drive train components of WT2. The red dashed line represents our set threshold at 0.06, the maximal reconstruction error before a segment is classified to be anomalous. Red shaded areas represent periods when the alarm notification criterion is active following three consecutive crossings of the threshold.

We applied a rule to trigger a fault alarm if three consecutive anomalies occurred, as indicated by the shaded areas in Figure 9. No alarm was triggered for the components associated with S1-S6. A small number of spectrograms from other components exceed the threshold of 0.5 according to the Isolation Forest models, as shown by sensors S2 and S4-S6 in Figure 8 but these did not result in persistently abnormal health scores, as shown in Figure 9. On the other hand, multiple persistent fault alarms were triggered for the generator coupling side towards the gearbox (S7, S8).

After detecting the fault-affected vibration responses from sensors S7 and S8, a consultation of the WT2 operation logs confirmed that the generator coupling side towards the gearbox had suffered incipient damage. Thus, both presented fault detection methods successfully detect a fault in the generator from the spectral features learnt by the convolutional autoencoder.

Our study makes use of multiple model parameter values, such as the bottleneck size of the autoencoder, the number of trees in the isolation forests, or the number of maximal consecutive days with anomalies in the health index. We optimized the parameter values related to the autoencoder architecture and training, specifically the number of layers, their sizes, and learning parameters, based on a hyperparameter optimization approach as described above. Optimizing these values is possible because the autoencoder training can be considered a supervised task in which the input spectrograms are taken to be the target outputs. Therefore, a loss value can be optimized by the choice of the autoencoder architecture and network parameters in the training process. On the other hand, the fault detection methods based on the spectrogram reconstruction error and based on the isolation forest perform unsupervised tasks because of the absence of a comprehensive set of labelled fault observations.

The optimal choice of parameters in the fault detection models is not in the scope of this study. Specifically, this includes the optimal number of trees in the isolation forest, the optimal contamination parameter in the isolation forest model, and the optimal reconstruction error threshold. We propose future studies to investigate in more detail the optimal choice of the parameter values based on additional fault observation datasets.

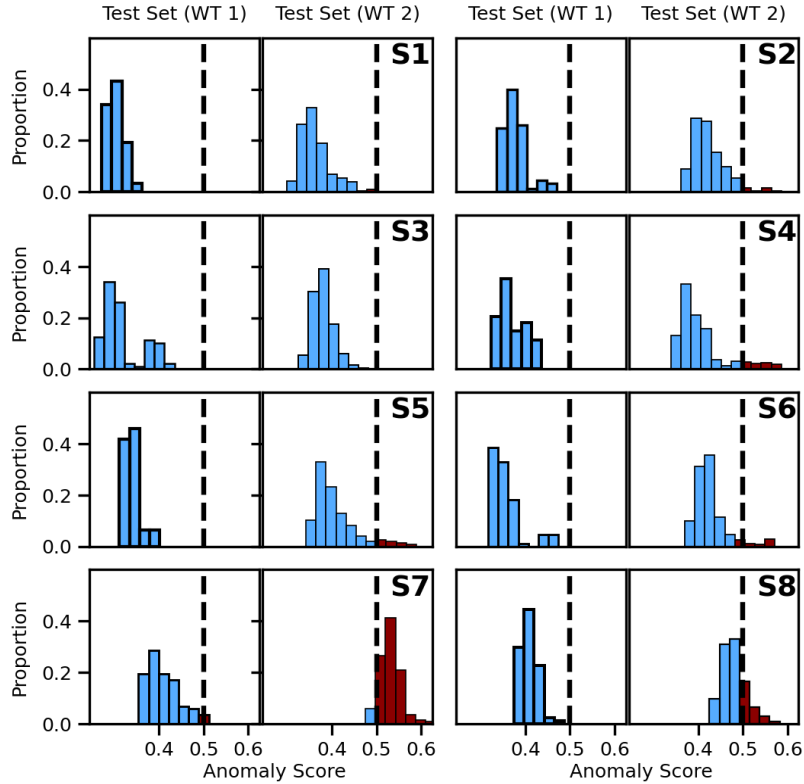

**Figure 8.** Fault detection based on isolation forest models. Visualization of the isolation forest anomaly scores for accelerometers S1-S8. In all cases, the vibration responses from WT1 are correctly classified as healthy based on their low anomaly score. Significantly larger anomaly scores are obtained for accelerometers S7 and S8 of WT2. This indicates a fault in the generator, which is consistent with the results of our reconstruction-error-based fault detection method (Figures 6 and 7).

We also investigated the effectiveness of the presented fault detection methods for detecting known gearbox damages in one of the two 750 kW turbines from NREL (section 3.1). Applying the same procedures as outlined for the commercial wind turbines above, we find that both fault detection methods are successful in detecting the fault-affected components in the damaged NREL wind turbine from their vibration responses. Our results from the NREL wind turbines confirm the effectiveness of both the reconstruction-based and the isolation-forest-based fault detection approach.

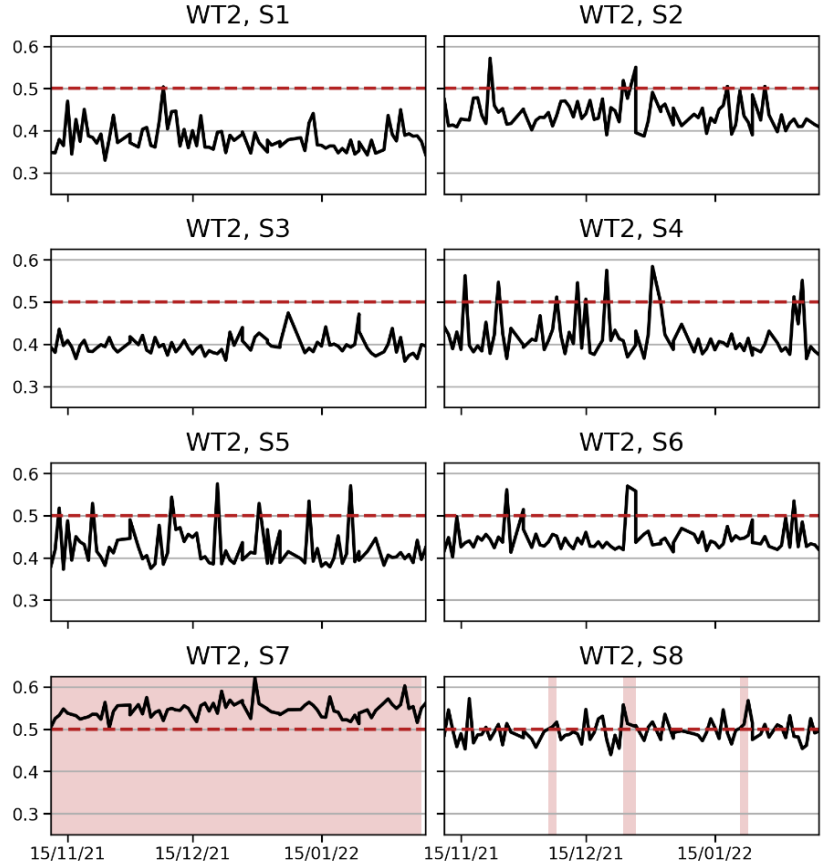

**Figure 9.** Fault detection based on isolation forest models. Health indices for the drivetrain components monitored by accelerometers S1-S8 of WT2. The red dashed line at 0.5 is used to separate normal vibration responses (health index ≲ 0.5) from anomalous ones (health index > 0.5). We set the threshold value of 0.5 in accordance with Liu et al., 2008. Red shaded areas represent periods when the alarm notification criterion is active following three consecutive crossings of the threshold.

## 5. Conclusions

Wind energy continues to expand strongly in Europe and around the world. The operation and maintenance costs of wind turbines account for a major fraction of the levelized cost of energy. Condition monitoring and artificial intelligence constitute powerful means for automating the early detection of incipient damage in wind turbines under various operation conditions. Machine learning methods enable early notification of wind turbine operators based on the vibration responses of the monitored components.

However, the existing vibration-based fault detection methods rely on the upfront definition of features in the frequency or time domains. This study has introduced two new fault detection methods for vibration-monitored parts that do not require any feature engineering. The proposed methods make use of convolutional autoencoders to detect unusual operation behavior from the spectrograms of the monitored parts. The autoencoders learn and extract spectral features customized to the monitored components in an autonomous manner without requiring any human assistance. In doing so, the autoencoders learn a spectral model of the component's normal behavior from past accelerometer measurements.

We have demonstrated the new fault detection approaches - based on reconstruction errors and based on isolation forests - in four wind turbine drive trains. We showed that both methods can successfully distinguish damaged from healthy vibration-monitored parts. First, we demonstrated their performances in detecting incipient generator damage from the vibration responses of the generators in two multi-MW onshore wind turbines. In addition, we confirmed the effectiveness of our presented fault detection methods in test rig measurements from NREL where they successfully detected gearbox damage.

Comparing convolutional and dense autoencoders for feature extraction and reconstruction, we found that convolutional autoencoders can accomplish the spectrogram reconstructions with a drastically lower number of parameters as compared to dense autoencoders. Thus, the proposed convolutional autoencoders avoid overfitting and can generalize better to unseen data.

Importantly, both presented fault detection methods do not require any feature engineering. We discussed how convolutional autoencoders can autonomously extract the most relevant features from the half spectrum. In this way, the autoencoders learn the normal vibration responses without requiring any upfront definition of features, thereby saving time and effort. An additional advantage of the presented methods is that the entire half spectrum can be monitored instead of the usual focusing on individual frequencies and harmonics.

The obtained results are very promising. While we demonstrated the new fault detection approach by means of gearbox and generator vibration responses, it can in principle also be applied in the structural health monitoring of other subsystems such as at the wind turbine tower. Further studies are needed for applications beyond the drive train to investigate the effect of variable operation conditions and how to account for them in the preprocessing and feature extraction in applications beyond the drivetrain. In the present study, we account for the effects of variable operation conditions by investigating the vibration responses at constant generator and drive train loads. The requirement to measure the vibration responses at constant loads can be easily accomplished in practical applications. For example, time-slice vibration measurements can be triggered whenever a specified load condition is met, e.g., by triggering the acquisition system at a certain rotational speed of the generator.

In addition to applications in other subsystems, future research should also investigate the proposed approach with more comprehensive datasets of vibration responses from damaged drive train components. It would be worthwhile to apply and investigate it for different damage types and intensities. It will also be interesting to study the temporal development of the proposed health indices in view of progressively increasing damage.

## Acknowledgements

The authors wish to thank Dr. Shawn Sheng from the National Renewable Energy Laboratory for providing test bench data from gearbox accelerometer measurements analysed in this study. The authors gratefully acknowledge that the present study was supported by a grant from the Swiss innovation agency Innosuisse.